\documentclass[nohyperref]{article}

\usepackage{microtype}
\usepackage{graphicx}
\usepackage{subfigure}
\usepackage{booktabs} %
\usepackage{multirow}
\usepackage[algo2e]{algorithm2e} 
\usepackage{hyperref}

\usepackage{enumitem}

\newcommand{\ie}{\textit{i}.\textit{e}., }
\newcommand{\eg}{\textit{e}.\textit{g}. }

\usepackage[accepted]{icml2022}

\usepackage{amsmath}
\usepackage{amssymb}
\usepackage{mathtools}
\usepackage{amsthm}
\usepackage{graphicx}
\usepackage[capitalize,noabbrev]{cleveref}

\theoremstyle{plain}

\theoremstyle{definition}

\theoremstyle{remark}

\usepackage[textsize=tiny]{todonotes}

\icmltitlerunning{GSCLIP : A Framework for Explaining Distribution Shifts in Natural Language}

\begin{document}

\twocolumn[
\icmltitle{GSCLIP : A Framework for Explaining Distribution Shifts in Natural Language}

\begin{icmlauthorlist}
\icmlauthor{Zhiying Zhu}{1}
\icmlauthor{Weixin Liang}{2}
\icmlauthor{James Zou}{2,3,4}
\end{icmlauthorlist}

\icmlaffiliation{1}{Department of Machine Learning, Carnegie Mellon University, PA, USA}
\icmlaffiliation{2}{Department of Computer Science, Stanford University, CA, USA}
\icmlaffiliation{3}{Department of Biomedical Data Science, Stanford University, CA, USA}
\icmlaffiliation{4}{Chan Zuckerberg Biohub, San Francisco, CA, USA}
\icmlcorrespondingauthor{James Zou}{jamesz@stanford.edu}

\icmlkeywords{Machine Learning, ICML}

\vskip 0.3in
]

\printAffiliationsAndNotice{}  %

\begin{abstract}

Helping end users comprehend the abstract distribution shifts can greatly facilitate AI deployment. Motivated by this, we propose a novel task, dataset explanation. Given two image data sets, dataset explanation aims to automatically point out their dataset-level distribution shifts with natural language. Current techniques for monitoring distribution shifts provide inadequate information to understand datasets with the goal of improving data quality. Therefore, we introduce GSCLIP, a training-free framework to solve the dataset explanation task. In GSCLIP, we propose the selector as the first quantitative evaluation method to identify explanations that are proper to
 summarize dataset shifts. Furthermore, we leverage this selector to demonstrate the superiority of a generator based on language model generation. Systematic evaluation on natural data shift verifies that GSCLIP, a combined system of a hybrid generator group and an efficient selector is not only easy-to-use but also powerful for dataset explanation at scale.
\end{abstract}

\section{Introduction}

A frequent cause for prediction failures in Machine Learning (ML) systems is the shifts between training and testing sets. Developing an understanding of these shifts is critical for successful AI design and deployment.

To better characterize these underlying information of datasets, we introduce a novel task, dataset explanation, to explain the distribution shifts between two data sets automatically with natural language. 
The shifts can vary from object classes, attributes, contexts ~\cite{oksuz2019imbalance, buda2017systematic, liang2022metashift, choi2012context}, subpopulation shift~\cite{gulrajani2020search}, to image-level concepts~\cite{peng2019moment, hendrycks2019benchmarking}. Dataset explanation can be extensively applied to downstream tasks including annotating the distribution shift benchmarks, evaluating dataset bias and limitations, or discovering model errors~\cite{eyuboglu2022domino,abid2021meaningfully},  which are indispensable for data-centric AI.

In mainstream literature\footnote{Related works are presented in Appendix~\ref{app:related_work}}, dataset explanation is primarily performed by statistical procedures for a binary question: whether there is a statistically significant shift or not~\cite{rabanser2019failing, raeder2009model}. 
These methods provide little insight into the structure of the distribution shift in a human-understandable way, \ie what is the shift. There are also no actionable instructions on what further training data to prioritize to collect for patching the model.

To tackle these challenges and generate open-ended and instructive dataset-level explanations, we propose a training-free workflow, GSCLIP. Since data can shift from disparate perspectives, the generator group should be able to produce a bunch of candidate explanations, and then the selector will quantitatively evaluate the candidates and pick up the most reasonable explanations. For this framework, the performance of the selector is crucial. We design a method based on CLIP~\cite{radford2021learning} and use a simple rule-based generator to validate the effectiveness of our proposed selector. To obtain more diverse and informative explanations, we introduce a powerful generator based on the large-scale pretrained language model~\cite{radford2019language, devlin2018bert}. We use the selector to further verify the effectiveness of the LM-based generator. Hence, our generator group incorporates two parts, the rule-based generator and the deep learning-based generator. In conjunction with an imaginative generator group and a powerful selector, we design an automated pipeline capable of explaining large-scale datasets without any training.
Experiments demonstrate that this workflow can provide not only open-ended but also precise explanations to AI users.

Our main contributions are summarized as follows: 
\begin{itemize}[leftmargin=*,nolistsep]
\itemsep0.5em
    \item We propose the first training-free and automated framework GSCLIP to solve dataset explanation. GSCLIP can propose a variety of coherent shift explanations in natural language and quantitatively evaluate them at scale.
    \item We leverage GSCLIP to demonstrate that the geometrics of a powerful class of recently-developed cross-modal embeddings can be used to summarize collected features of a data set.
    \item GSCLIP can be directly applied to assist AI deployment for data-centric AI, \eg model error discovery, and subgroup bias detection. 

\end{itemize}

\section{Methodolody}

\subsection{The GSCLIP Pipeline}
In this section, we introduce our GSCLIP system.
First, given two image datasets $\mathcal{D}^\text{A}$ and $\mathcal{D}^\text{B}$, the candidate generator outputs a corpus of candidate explanations $\mathcal{S}_\text{all}$. Second, the selector will quantitatively evaluate them and pick the most salient and reasonable ones as answers. Since distribution shifts can be multimodal when translated into natural language(\eg shifts between a set of black outdoor dogs and a set of white indoor cats), our workflow is designed to reveal a \textbf{variety} of \textbf{feasible} answers to end users. 

\subsection{Candidate Generator}
\label{meth:generator}
The candidate generator group consists of a rule-based generator and a LM-based generator. The two separate generators produce diverse and human-understandable natural language descriptions of distribution shifts, using certain phrase templates, which vary from tasks and test data sets. The templates are inspired by recent study on prompt learning for large pre-trained models~\cite{radford2021learning, zhou2021learning}. More implementation details about the templates are placed in In Section \ref{sec:exp}.

\paragraph{Rule-Based Generator}
We start from constructing natural explanations by rules. To procure the rule-based explanation corpus $\mathcal{S}_\text{rule} = \{t_i\}_{i=1}^{n_\text{1}}$, the rule-based generator fills the blank slots of pre-defined templates with textual annotations from image datasets. \eg For CelebA dataset~\cite{liu2015faceattributes}, the generator maps the ground truth category labels to attribute descriptions, fill up the template "a photo of a [slot] with [slot]" with selected attributes(male, black hair etc.), and get the candidate explanation "a photo of a male with black hair". We assume that "a photo of a male with black hair" and "a photo of a male with blond hair" are both feasible explanations of the shifts between photos of black-haired men and photos of blonde-haired men. Therefore, one benefit of utilizing the rule-based generator is to ensure that there are viable answers in $\mathcal{D}_\text{all}$.
\begin{figure*}[!htb]
	\centering
	\includegraphics[width=0.99\textwidth]{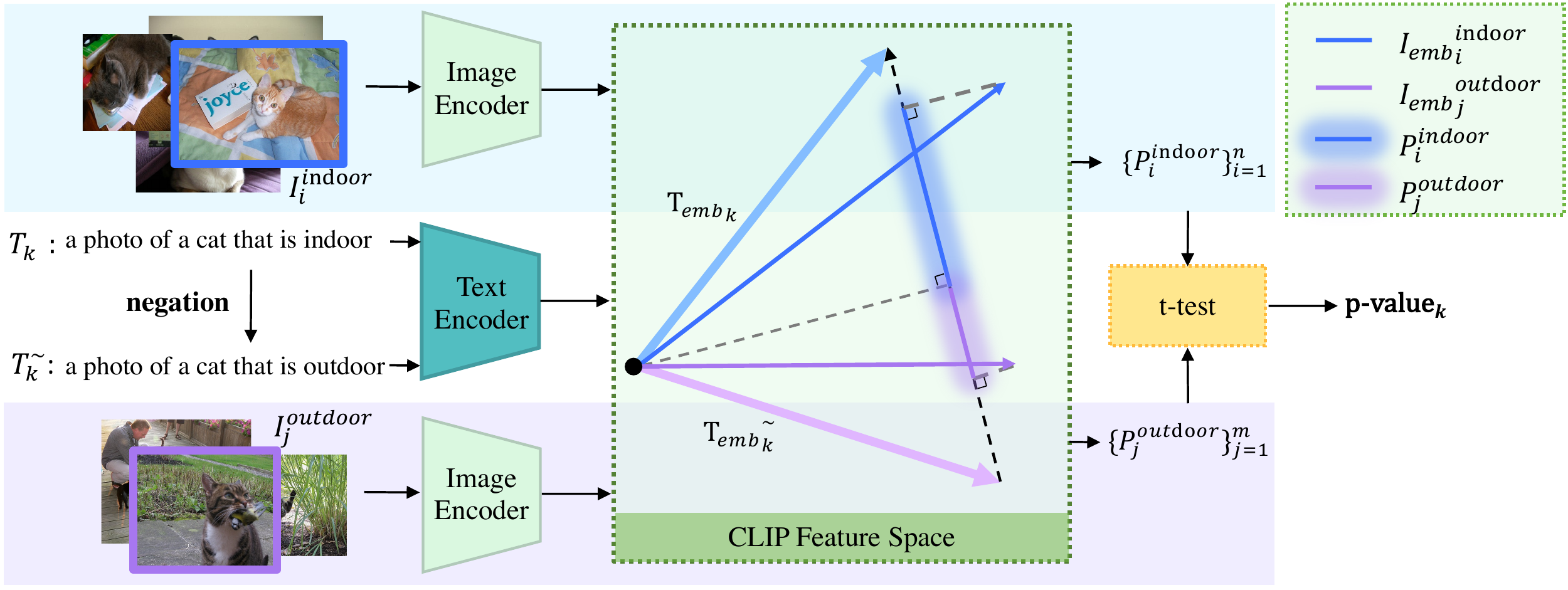}
	\caption{The explanation selector in our GSCLIP workflow. For demonstration purposes, we present the high-dimensional CLIP space in a 2-D drawing. ${T}_\text{k}, {T}_\text{k}^{\sim}$ denote the $k_{th}$ candidate sentence and its negation in the "cat" subset of our curated corpus, and  ${T_{emb}}_\text{k}, {T_{emb}}_\text{k}^{\sim}$ are their embeddings. Similarly, ${I}_\text{i}^{indoor}, {I}_\text{j}^{outdoor}$ means the $i_{th}$ and $j_{th}$ image in the two test image sets respectively, which are encoded as ${I_{emb}}_\text{i}^{indoor}, {I_{emb}}_\text{j}^{outdoor}$, and their projections are represented as ${P}_\text{i}^{indoor}$ and ${P}_\text{j}^{outdoor}$. After traversing test image sets, we derive $\{{P}_\text{i}^{indoor}\}_{i=1}^{n}$, $\{{P}_\text{j}^{outdoor}\}_{j=1}^m$. Note that $n$ and $m$ are not necessarily equal.}
	\label{fig:method_selector}  
\end{figure*}

\paragraph{LM-Based Generator}
Beyond the rule-based generator, we want a more powerful generator to enrich $\mathcal{D}_\text{all}$ with more diverse, flexible and informative sentences for systematic distribution shift evaluation. We introduce pre-trained generative LM to obtain a coherent text set $\mathcal{S}_\text{LM} = \{t_j\}_{j=1}^{n_\text{2}}$. Specifically, we use sentences such as "a photo of a cat that is indoor" that start with the uniform form of "a photo of \ ($<$article$>$) $<$object$>$ $<$preposition$>$" or "a photo of \ ($<$article$>$) $<$object$>$ that", since preliminary experiments show that sentence patterns have little effect on the performance of our framework. For a specific experiment, we choose one form(and $<$preposition$>$) of the sentences, and fill the $<$object$>$ slot with the object class names in the data set (\eg MetaShift~\cite{liang2022metashift}, ImageNet~\cite{deng2009imagenet}) vocabulary to derive sentence prefixes. Next, we use GPT-2~\cite{radford2019language} to find all completions of these sentence prefixes over a certain probability threshold and get a list of candidate sentences sorted by their probabilities in descending order. It's verified by further experiments that the deep-learning-based method can generate imaginative and coherent explanations while keeping the inference accuracy.

In our final pipeline, we include both rule-based generator and LM-based generator in the hybrid generator group and get $\mathcal{S}_\text{all} = \mathcal{S}_\text{rule} + \mathcal{S}_\text{LM}$.

\subsection{Explanation Selector}
\label{meth:selector}
The performance of the GSCLIP framework largely hinges upon the way of prioritizing candidate explanations. However, no off-the-shelf method provides automated and quantitative evaluation for measuring the distribution shift explanation. As shown in Figure~\ref{fig:method_selector}, we introduce the explanation selector for large-scale, systematic and quantitative explanation evaluation and selection.

The proposed explanation selector follows 4 steps: 
\begin{enumerate}[leftmargin=*,nolistsep]
\itemsep0.6em
  \item Get sentence pairs by negation. Since we are more interested in object attributes, co-occurrence, etc., we fix the main object $\mathcal{O}^\text{main}$  (\eg cat) and only test on relevant image sets and candidate explanation set for simplicity. For each candidate explanation,  we fix $\mathcal{O}_\text{main}$ and negate the descriptive part of the sentence, \eg "a photo of a cat with grass" $vs$ "a photo of a cat without grass". An alternative to the negation method is to pair the candidate explanation with a general statement, \eg "a photo of a cat with grass" $vs$ "a photo of a cat". Extensive experiments show that the two methods achieve comparable performance.
  
  \item Encode image sets and candidate explanation set relevant to $\mathcal{O}^\text{main}$ into multi-modal embedding space. We adopt the cross-modal pre-trained model CLIP to embed images and sentences in the same high-dimensional representation space. Note that the textual features can be cached in advance for fast retrieval in this step. We denote the embedding collection of  $\mathcal{D}^\text{A}$, $\mathcal{D}^\text{B}$ as $\mathcal{D}_{emb}^\text{A}$, $\mathcal{D}_{emb}^\text{B}$. %

  \item Calculate vector projections. Motivated by the geometrical features of high-dimensional vectors, we propose a novel projection method to quantitatively measure the relevance between the hypothetical and the actual shift. For each visual vector $E_{emb}^{A_{i}}, i=1,2,...n$ in $\mathcal{D}_{emb}^\text{A}$, we 
  define the projection as ${P}_{k}^{A_{i}} = \ \mid E_{emb}^{A_{i}} \cdot Diff_{k} \mid $, where $Diff_k$ is the difference vector between sentence embedding ${T_{emb}}_{k}$ and its negation ${T_{emb}}^{\sim}_{k}$. After traversing $\mathcal{D}_{emb}^\text{A}$, we obtain a list of projection $L_{A} =  \{P_k^{A_{i}}\}_{i=1}^{n}$. Similarly, we derive $L_{B} =  \{P_k^{B_{j}}\}_{j=1}^{m}$ from $\mathcal{D}_\text{B}$. The pseudo-code for the algorithm is presented in Appendix~\ref{app:pseudocode}.

  \item Conduct t-test. We conduct a two-sample t-test on $L_{A}$ and $L_{B}$ and quantitatively assess the natural explanation by p-value. If p-value \textless \ 0.05, we consider $L_{A}$ and $L_{B}$ are statistically different as the $k_{th}$ candidate explanation states. Last, We sort candidate sentences by p-value in ascending order and get the top-x reasonable explanations.
\end{enumerate}
The explanation selector takes different directions into consideration to validate each candidate explanation. By running the selector, an accurate, coherent and comprehensive explanation set is produced.

\section{Experiments}
\label{sec:exp}

\subsection{Experimental Setup}

\paragraph{Datasets}
For systematic and large-scale inference using our GSCLIP framework, we experiment with the recently introduced MetaShift and MetaShift-Attributes datasets \footnote{https://github.com/Weixin-Liang/MetaShift}. Different to previous benchmark datasets related to distribution shifts ~\cite{gulrajani2020search,sagawa2019distributionally, venkateswara2017deep, guo2020broader}, MetaShift contains comprehensive cases of realistic dataset-level shifts(12,868 sets of natural images across 410 classes) with explicit annotations. We utilize MetaShift to monitor shifts on object co-occurence and MetaShift-Attributes to monitor object attributes.

\paragraph{Implement Details}
For model architecture, we adopt CLIP\footnote{https://github.com/moein-shariatnia/OpenAI-CLIP} with a  "ViT-B/32"~\cite{dosovitskiy2020image} image encoder as the backbone of our selector. In the experiments below, we randomly sample 100 data set pairs from MetaShift (or MetaShift-Attributes) for systematic evaluation. Note that we are especially interested in monitoring the more fine-grained shift, which is also more challenging. Therefore, for each test data set pair, we add the constraint during sampling that their main object $\mathcal{O}^\text{main}$ should be consistent(\eg The two sets are all cat images). We conduct the inference on a single 16G Nvidia V100. 

\paragraph{Performance Evaluation}
We conduct a quantitative evaluation of our GSCLIP pipeline. With the ranking over hypotheses returned by our framework, we can calculate the top-x accuracy by checking the ground truth labels in the explanation hypotheses generated by rules. The explanation sentence is deemed accurate once it contains the annotation word of any of the test image sets. Under the setting when multiple selectors are jointly used, we also 
consider a $Key\ Word$ metric, namely taking both the ground truth labels and their WordNet synonyms~\cite{miller1998wordnet} into consideration when computing the top-x accuracy.

\subsection{Systematic Evaluation of GSCLIP}
\label{exp:sys}

\subsubsection{The Effectiveness of The Selector}

To validate the selector's ability to detect shifts, we conduct a large-scale evaluation with only the rule-based generator.  
The setting is similar to a k-class image classification problem, where k represents the number of rule-based hypotheses for each object. The mean of k is 66 for MetaShift and 14 for MetaShift-Attributes.

We list the results in Table~\ref{tab:selector}. We can see that the selector achieves promising results with only the rule-based generator. There is a high probability for the selector to pick up the ground truth explanations from candidates. That is to say, the selector is capable of explaining shifts related to object co-occurrence and object attributes.

\begin{table}[htbp]
  \centering
    \begin{tabular}{c|c|c|c}
     \toprule
    \multirow{2}[1]{*}{Dataset} & \multicolumn{3}{c}{Accuracy} \\
          & \multicolumn{1}{c}{Acc@1} & \multicolumn{1}{c}{Acc@3} & \multicolumn{1}{c}{Acc@5} \\
    \midrule
    MataShift & 30\%  & 50\%  & 63\% \\
    \midrule
    MetaShift-Attributes & 34\%  & 57\%  & 71\% \\
    \bottomrule
    \end{tabular}%
     \caption{The top-x accuracy (denoted as Acc@x in the table) in the systematic evaluation on rule-based explanations. Note that in this experiment we set $x < 5$ to ensure the candidate number is large than x.}
  \label{tab:selector}%
\end{table}%

\subsubsection{the effectiveness of LM-based generator}
Another concern for GSCLIP is that the rule-based generator only produced limited types of shift explanations, which are not imaginative and comprehensive enough for real-world applications. We design the following experiments to demonstrate that introducing the LM-based generator makes our generator more powerful. Additionally, we try an alternative generator -- a programmatic approach that creates candidate explanations from
most frequently used words on English Wikipedia ~\cite{semenovWikiRepo} with part-of-speech (POS)~\cite{petrov2011universal} constraints.
Due to the GPU memory limitation, we use the top-3700 candidates related to $\mathcal{O}^\text{main}$ with the highest generation possibility for the LM-based selector, and with the highest frequency for the Wikipedia baseline. Together with candidate explanations generated by rule-based techniques, we can derive thousands of candidate explanations for each $\mathcal{O}^\text{main}$.

The quantitative results on MetaShift and MetaShift-Attributes are listed in Table \ref{tab:res_all}. We can see that 1) the LM-based generator accomplishes the candidate explanation generation improvement by a large margin, which accounts for the 4\% - 13\% accuracy increase when comparing $Key\ Words$ to $Label$; 2) the LM-based generator is more powerful than the rule-based one, because Acc@1 and Acc@5 of $Label$ in $GPT-2$ in Table~\ref{tab:res_all} is lower than that of Table~\ref{tab:selector}, which means the explanations generated by GPT-2 rank higher than those generated by rules; 3) the LM-based generator largely surpasses the Wikipedia baseline, since pre-trained LM also consider semantics and sentence coherency besides word frequency when generating texts; 4) GSCLIP achieves promising accuracy in the challenging data explanation task, which proves the superiority of our proposed pipeline.

\begin{table}[htbp]
  \centering
  \resizebox{0.5\textwidth}{!}{%
    \begin{tabular}{c|c|c|c|c|c}
    \toprule
    \multirow{2}[0]{*}{Dataset} &       &       & \multicolumn{3}{c}{Accuracy} \\
          &       &       & \multicolumn{1}{c}{Acc@1}  & \multicolumn{1}{c}{Acc@5}  & \multicolumn{1}{c}{Acc@15} \\
    \midrule
    \multirow{4}[0]{*}{MetaShift} & \multirow{2}[0]{*}{GPT-2} & Label    & 28\%  & 46\%  & 55\% \\
    \cmidrule{3-6}
          &       & Key Words & 39\%  & 54\%  & 64\% \\
          \cmidrule{2-6}
          & \multirow{2}[0]{*}{Wiki} & Label    & 26\%  & 38\%  & 43\% \\
          \cmidrule{3-6}
          &       & Key Words& 28\%  & 42\%  & 48\% \\
    \midrule
    \multirow{4}[0]{*}{Meta-Attr} & \multirow{2}[0]{*}{GPT-2} & Label    & 20\%  & 29\%  & 36\% \\
    \cmidrule{3-6}
          &       & Key Words & 25\%  & 40\%  & 49\% \\
          \cmidrule{2-6}
          & \multirow{2}[0]{*}{Wiki} & Label    & 11\%  & 15\%  & 22\% \\
          \cmidrule{3-6}
          &       & Key Words & 16\%  & 22\%  & 32\% \\
    \bottomrule
    \end{tabular}%
    }
    \caption{The comparison of different generators and evaluation metrics for shift explanation. $GPT-2$ means the generator group consists of the rule-based generator and the LM-based one. $Wiki$ represents the rule-based generator + Wikipedia-based baseline. $Label$ means explanations are considered correct only if they contains ground truth label words. In the case of $Key\ Words$, both the ground truth labels and their synonyms are taken into account.}
  \label{tab:res_all}%
\end{table}%

\section{Conclusion}
In this work, we propose GSCLIP, the first automated workflow for the image dataset explanation task, which is based on a hybrid generator group and a powerful selector. We leverage a pretrained language model to allow the generation of coherent and imaginative candidate explanations. And we utilize the geometrical features of high-dimensional cross-modal embeddings to quantitatively evaluate each shift explanation. The whole system is general, training-free and can scale up to large datasets. Systematic evaluations show that GSCLIP accomplishes the dataset explanation task.

GSCLIP can be directly applied to downstream tasks in data-centric AI from design to post-deployment. For example, we can identify and remove bias and limitations of a dataset by comparing it to a balanced set (\eg GQA~\cite{hudson2019gqa}), and can audit model errors by analyzing the difference between training and inference sets or subgroups within the training set.

\bibliography{example_paper}
\bibliographystyle{icml2022}

\newpage
\appendix
\onecolumn
\section{Related Work}
\label{app:related_work}
\subsection{Monitoring Distribution Shift}
ML models can have divergent behaviors when distribution shift happens. To characterize distribution shift, \citet{schneider2020datashiftexplorer} leverages visualization technique to identify, analyze, and compare the
change in multidimensional data distributions, \citet{rabanser2019failing} and \citet{feutry2019simple} perform it by statistical procedures, \citet{ginart2022mldemon} models a human-in-the-loop approach to minimize the number of required verifications to detect dataset shift. However, these approaches primarily monitor distribution shift by transforming it into a binary question: whether there is a significant shift or not. This provides little information about what is the shift in a human-understandable way. Besides, there are few insights about how to remove these shifts for better data quality and better AI deployment. Recently, we notice a concurrent work~\cite{zhong2022summarizing} using a similar proposer and verifier network on single text modality dataset, but the system needs 
training procedure of the neural networks.

\subsection{Benchmark Datasets for Distribution Shifts}
Existing image benchmarks for dataset shift can be dichotomized into dataset of synthetic distribution shifts and real shifts.

Synthetic shift datasets are mainly obtained by image transformation, \eg ImageNet-C, transformed MNIST and CIFAR, DomainNet~\cite{hendrycks2019benchmarking, worrall2017harmonic, peng2019moment}. 
These transformations are clearly defined for supervising distribution shift. However, they cannot represent natural shifts and are insufficient for real-world applications. Datasets about real-world shift are usually collected from various experiments or sources. Unfortunately, they either lack systematic annotations or have limited scales.

Recently, \citet{liang2022metashift} present MetaShift—a large scale benchmark dataset for distribution shifts. MetaShift also provides explicit annotations of the differences between any two sub-datasets, and all these features lay the basis for much more comprehensive assessment of distribution shifts.

\section{Pseudo-code}
\label{app:pseudocode}
\begin{algorithm}[!h]
\caption{The projection method of the selector in GSCLIP.}
\label{alg}
  \SetKwInOut{Input}{Input} %
    \Input{A text embedding set $\{{T_{emb}}_{k}\}_{k=1}^{l}$ and  $\{{T_{emb}}^{\sim}_{k}\}_{k=1}^{l}$; Two image embedding sets $\{{I_{emb}}_{i}^A\}_{i=1}^{n}$ and $\{{I_{emb}}_{j}^B\}_{j=1}^{m}$.}
    \SetKwInOut{Output}{Output} %
     \Output{A set of p-value $\{{p-value}_{k}\}_{k=1}^{l}$}
     
    \For{$k=1, 2, ..., l$}{
        $Diff_k$ = ${T_{emb}}_{k}$ - ${T_{emb}}^{\sim}_{k}$
        
        \For{$i=1, 2, ..., n$}{
        ${P}_{k}^{A_{i}} = \ \mid {I_{emb}}_{i}^A \cdot Diff_{k} \mid $
        }
        
        \For{$j=1, 2, ..., m$}{
        ${P}_{k}^{B_{j}} = \ \mid {I_{emb}}_{j}^B \cdot Diff_{k} \mid $
        }
        
        ${p-value}_{k} = t-test(\{{P}_{k}^{A_{i}}\}_{i=1}^n,       \{{P}_{k}^{B_{j}}\}_{j=1}^m)$
        
     }
    
\end{algorithm}

\end{document}